\documentclass[10pt,twocolumn,letterpaper]{article}

\usepackage{cvpr}
\usepackage{times}
\usepackage{epsfig}
\usepackage{graphicx}
\usepackage{amsmath}
\usepackage{amssymb}
\usepackage{subfigure}

\usepackage[pagebackref=true,breaklinks=true,letterpaper=true,colorlinks,bookmarks=false]{hyperref}

\cvprfinalcopy 

\ifcvprfinal\fi
\begin{document}

\title{Generating Attention from Classifier Activations for Fine-grained Recognition}

\author{Wei Shen \qquad Rujie Liu \\
Fujitsu Research \& Development Center, Beijing, China.
\\
{\tt\small \{shenwei, rjliu\}@cn.fujitsu.com}
}

\maketitle

\begin{abstract}
   Recent advances in fine-grained recognition utilize attention maps to localize objects of interest. Although there are many ways to generate attention maps, most of them rely on sophisticated loss functions or complex training processes.    
   In this work, we propose a simple and straightforward attention generation model based on the output activations of classifiers. The advantage of our model is that it can be easily trained with image level labels and softmax loss functions. More specifically, multiple linear local classifiers are firstly adopted to perform fine-grained classification at each location of high level CNN feature maps. The attention map is generated by aggregating and max-pooling the output activations. Then the attention map serves as a surrogate target object mask to train those local classifiers, similar to training models for semantic segmentation.
   Our model achieves state-of-the-art results on three heavily benchmarked datasets, \ie 87.9\% on CUB-200-2011 dataset, 94.1\% on Stanford Cars dataset and 92.1\% on FGVC-Aircraft dataset, demonstrating its effectiveness on fine-grained recognition tasks.
\end{abstract}

\section{Introduction}

\noindent Fine-grained recognition aims at distinguishing subordinate categories (e.g., bird species~\cite{wei2018mask,branson2014bird,lin2015bilinear}, dog species~\cite{simon2015neural,44891,zhao2016diversified,Niu_2018_CVPR}, car models~\cite{krause2015fine,zheng2017learning,fu2017look,Cui_2018_CVPR}, etc.). 
The challenge is that visual differences between subordinate categories can be subtle while the variation within the same category can be very large.

In response to the challenge, the first step of most fine-grained recognition models is to localize the object region
Models are usually trained with ground truth structured annotations (e.g., part annotations, bounding boxes, attributes, etc.)~\cite{branson2014bird,krause2015fine,krause2014learning,zhang2014part,lin2015deep,wang2016mining,huang2016part,zhang2016spda,liu2017localizing}. 
However, those annotations require either domain knowledge or extensive crowd sourcing which are expensive.
Recently, visual attention mechanism has been introduced to many image-related tasks~\cite{zhang2016top,gregor2015draw,xu2015show,yang2016stacked,chen2016attention,xu2016ask,wang2017residual,wang2017multi}. In the scenario of fine-grained recognition, it helps the recognition model focus on discriminative regions~\cite{44891,zhao2016diversified,zheng2017learning,fu2017look,liu2017localizing,xiao2015application}. 
In general, visual attention can be formulated in either a bottom-up manner or a top-down manner~\cite{connor2004visual}. In the bottom-up manner, the region attention is generated by performing unsupervised clustering on the neural activations~\cite{simon2015neural,zheng2017learning,xiao2015application,zhang2016picking,wang2015multiple}. The basic idea is that some high level neurons can only be activated by similar image patterns. 
By analyzing the concurrence of the activations of those neurons and object parts, one can build multiple part localizers which will be used for generating part proposals in the testing phase. 
However, splitting the attention localization apart from the recognition task may yield a sub-optimal solution to the recognition task.

In contrast, the top-down region attention is learned to minimize the recognition loss in an end-to-end way. Nevertheless, generating accurate attention is quite challenging since less prior knowledge is involved and the image level labels are too coarse to generate attention maps. In order to learn region attention that is helpful for the final recognition task, researchers have to devise auxiliary objective functions and specialized optimization strategies to regularize the attention learning process so that the obtained attention is category-discriminative, compact and diverse. However, those loss functions and optimization strategies add much complexity to the model and may increase the difficulty of model training.

To the goal of generating effective attention regions in a simplified way (\ie without specially designed objective functions or training strategies), we propose to make use of the activations from classifiers (see Figure~\ref{fig:pipeline}). Those classifiers make fine-grained category predictions at each location on the high level feature maps and their output activation volumes are then aggregated to yield a single volume representation which is further pooled to generate the attention map. The benefits of our model is two-fold. (1) The entire generation part consists of only training multiple classification models which is quite standard. Therefore, there is no extra complex customized strategy and we can use softmax loss to train the whole model. (2) Our attention map is produced by classifiers other than unsupervised clustering. It ensures the attended region is category-discriminative. Hence, those regions can be used for further analysis in a multi-scale recognition way.

\section{Motivation and Contributions}
Our motivation is that if a local region can be used by a classifier to make a correct prediction of the object category, it means the region is discriminative and needs our predictor's attention. An attractive property of generating attention from classifier activations is that when we apply the classifier at each location we can obtain a probability distribution over all categories. Rather than a binary output that only indicates the possibility of being the attention region or not, this probability distribution allows us to obtain abundant information about the local region. If we have multiple classifiers, the aggregated predictions will further allow us to amend the incorrect predictions of each classifier through backpropgation during training. For example, given a local region, classifier A may predict it to be a background region with a modest probability, while classifier B may predict it to be one of the fine-grained category with a high probability. Then we will regard this local region as a discriminative region and train classifier A so that it can make correct target fine-grained prediction next time. This leads us to the idea of training each classifier using the attention map as the surrogate label (see Section~\ref{sec:learn_attn} for details). 

Our work is also inspired by ~\cite{zhou2016learning} in which the authors find that a simple modification of the pooling layer can allow the classification-trained CNN model to localize attention regions (or category-specific image regions) in a single forward-pass. In this work, instead of using the pooling layer to obtain attention regions, we adopt the activations from multiple classifiers, which contain  category-specific information, to extract attention regions. 

The contributions of our work can be summarized as follows.
\begin{itemize}
\item We propose a novel attention generation method for fine-grained recognition based on the output activations of multiple classifiers. The whole learning process is supervised under softmax loss functions with image level labels. Our model is much simpler compared to many state-of-the-art models.
\item We demonstrate that local classifiers can be trained in a way similar to training semantic segmentation models and the generated attention maps can well serve as a surrogate target object mask in the training process.  
\item We conduct extensive experiments on three heavily benchmarked datasets (\ie CUB-200-2011, Stanford Cars and FGVC-Aircraft datasets) and achieve new state-of-the-art recognition results.  
\end{itemize}

\section{Related Work}

Fine-grained recognition models usually contain object region localization as an important component. According to how models learn the object region, we categorize them into two categories: supervised localization based models and visual attention models.
Since models in the second category are more related to our work, we will give a brief introduction to the first category and more details of the second category.

\subsection{Supervised Localization Based Models}
A straightforward way to find the object region is to train models using object annotations~\cite{branson2014bird,zhang2014part,lin2015deep,wang2016mining,huang2016part}. 
Branson~\etal~\cite{branson2014bird} use the detected object keypoints to align multiple wrapped regions with prototypical models and extract region features using a deep convolutional neural network. 
Lin~\etal~\cite{lin2015deep} propose a valve linkage function to enable the back-propagation through fine-grained recognition, part alignment and localization, connecting all sub-networks. 
Wang~\etal~\cite{wang2016mining} mine triplets of patches with geometric constraints to automatically find discriminative regions which help improve the accuracy of recognition.

Models in this category usually require ground truth annotations during the training phase. However, collecting those annotations are time-consuming and labor-expensive, posing a problem for scaling up the models to large fine-grained datasets.

\begin{figure*}
\centering
\includegraphics[width=0.99\linewidth]{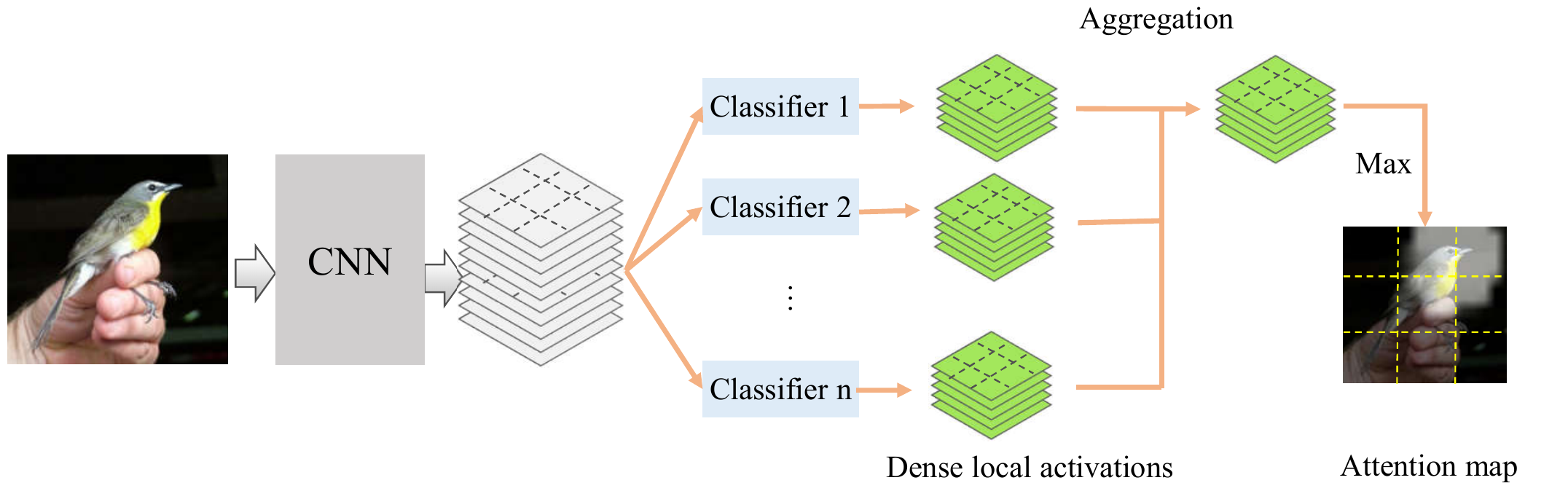} 
\caption{The pipeline of generating the attention map from classifier activations. Given an image, we firstly extract high level convolutional feature maps. Then $n$ local classifiers are applied at each location on the feature maps to produce $n$ corresponding dense local activation volumes. Each slice of the volume stores the attention map of the corresponding category. Those volumes are aggregated and max-pooled along the channel dimension to generate the final attention map. (Best viewed in color.)
}
\label{fig:pipeline}
\end{figure*}

\subsection{Visual Attention Models}
In order to overcome the weakness of the supervised localization based models, researchers resort to visual attention mechanism in either a bottom-up or top-down way~\cite{simon2015neural,zhao2016diversified,zheng2017learning,fu2017look,xiao2015application,zhang2016picking,wang2015multiple,jaderberg2015spatial}. 
\subsubsection{Bottom-up Visual Attention Models} 
In fine-grained recognition, bottom-up visual attention models usually perform neuron activation clustering to localize object parts. The basic idea is that some neurons can respond to specific image patterns significantly and consistently (\eg, some neurons are activated when they see bird's beaks)~\cite{zhang2016picking}. By analyzing those activations, one can detect object parts in an unsupervised way. 
Simon~\etal~\cite{simon2015neural} propose to find constellations from deep neural activation maps and learn a part model by selecting part detectors that fire at similar relative locations. 
Xiao~\etal~\cite{xiao2015application} perform spectral clustering to partition filters into several groups each of which acts as the part detector.
Zhang~\etal~\cite{zhang2016picking} introduce a two-step picking strategy in which the first step picks filters that respond to specific patterns to train object detectors and the second step picks filter responses via spatially weighted fisher vector encoding to obtain the final image representation.

Bottom-up visual attention models do not need part annotations. However, the attention sub-model is optimized independently of the fine-grained recognition task, which may lead to a sub-optimal solution.

\subsubsection{Top-down Visual Attention Models}
In order to obtain optimal visual attention for fine-grained recognition,
the top-down visual attention is learned so that it can help minimize the objective function of the fine-grained recognition.
Fu~\etal~\cite{fu2017look} propose a recurrent attention convolutional neural network which learns discriminative region attention from coarse to fine. 
Zhao~\etal~\cite{zhao2016diversified} introduce a diversified visual attention network that proposes multiple canvases for high level feature extraction and predicts the attention map for each canvas using attention LSTMs.
Pierre~\etal~\cite{44891} propose a recurrent neural network that can generate multiple “glimpses” into the input image and extracts multi-resolution patches for the final recognition.

Top-down visual attention models unify part localization and discriminative feature learning for fine-grained recognition as a whole. Nevertheless, researchers have to take special care (\eg pairwise ranking loss~\cite{fu2017look}, diversity loss~\cite{zhao2016diversified}) when designing top-down visual attention models. 

The most relevant work to ours is~\cite{fu2017look}, whereas there are noticeable differences in three aspects. (1) Our model is trained using purely softmax loss while ~\cite{fu2017look} adopts an additional ranking loss. (2) Our model generates object attention from the activations of multiple classifiers while ~\cite{fu2017look} generates attention from stacked fully-connected layers. (3) The attention in this work is in the form of two-dimensional masks while the attention in ~\cite{fu2017look} is represented by bounding boxes.

\section{Approach}
The pipeline of the our attention generation process is shown in Figure~\ref{fig:pipeline}. It firstly extracts high level feature maps from the image via a backbone CNN. Then multiple classifiers are applied on the feature maps to output category prediction activation volumes which are then aggregated to obtain the attention map. 
For fine-grained recognition tasks, we use our generated attention maps to crop and zoom the attended regions to perform multi-scale predictions. 
Details will be provided in the following subsections.

\subsection{High Level Feature Extraction}
We use a fully convolutional network (FCN) for high level feature extraction. The reason for adopting a FCN is two-fold. One is that the FCN outputs convolutional feature maps which can be used for attention map generation. The other is that the FCN can ideally handle input images with different sizes so that we do not need to crop images during test time. We denote the output of the FCN as $f(X)$, where $X$ is the input image. The size of $f(X)$ is $C\times H\times W$, where $C$ is the number of channels, $H$ and $W$ are the spatial height and width of the feature maps.

\begin{figure}
\centering
\includegraphics[width=0.99\linewidth]{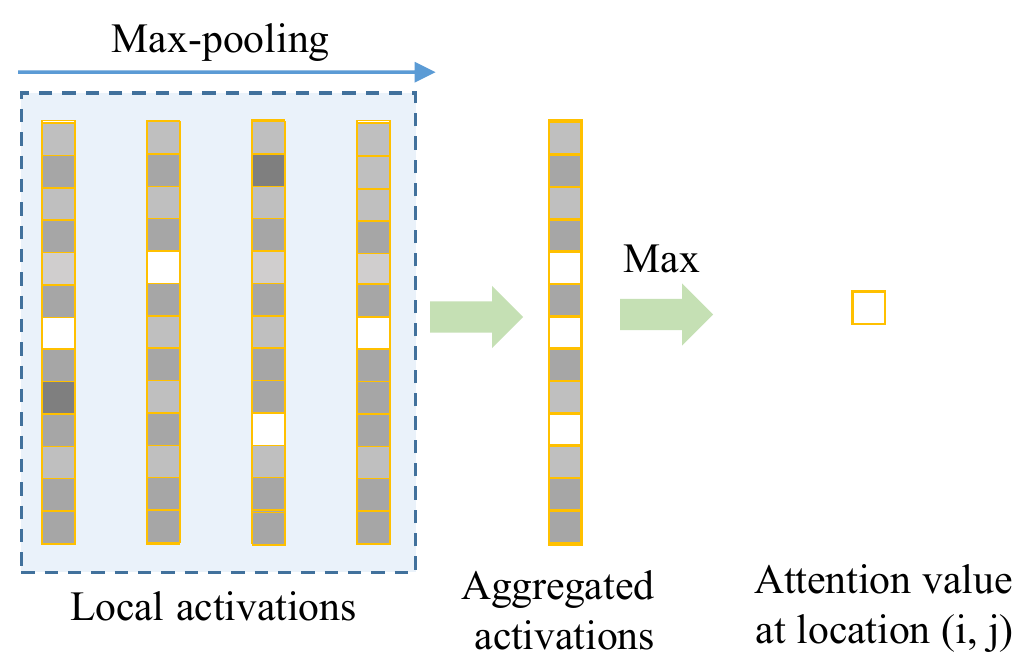} 
\caption{An example of generating the attention value at location (i, j) from the activations of four local classifiers. (i, j) stands for a location index on the feature maps. 
A brighter square means higher activation value indicating the prediction confidence of the classifier. We max-pool the activation vectors to obtain an aggregated version. It then is used to generate attention value at location (i, j) by simply max-pooling along the channel dimension excluding the last background category. We repeatedly perform this process at each location of the feature maps. 
}
\label{fig:get_attn}
\end{figure}

\subsection{Dense Local Activations}
Before the detailed description of the core components of our model, we firstly introduce concepts that will be used in this paper.

\noindent\textbf{Local classifiers}. In this work, local classifiers are linear classifiers that output $(L+1)$ dimensional activations. $L$ stands for the number of fine-grained categories. The extra one is used to indicate the background category. The input to those classifiers are the feature vectors along the channel dimension in the convolutional feature maps. 
Intuitively, performing classification based on those feature vectors is equivalent to performing classification based on local image patches since the receptive field of each high level neuron corresponds to a local region on the input image.
 

\noindent\textbf{Dense local activations}. Dense local activations are the output of local classifiers when they are applied at each location on the feature maps. More specifically, given the $C\times H\times W$ feature maps, we treat them as $H\times W$ feature vectors with dimension $C$. Each local classifier will classify those $H\times W$ feature vectors into $(L+1)$ categories, \ie either one of the fine-grained categories or the background category. 
Therefore, each classifier will output a $(L+1)\times H\times W$ activation volume which we call \textit{dense local activations}. Each slice of the volume is an attention map of the corresponding category including the background category.

Images are high dimensional data such that using a single local classifier may be insufficient to achieve high recognition accuracy. Therefore, we adopt $n$ local classifiers to obtain $n$ activation volumes $\textbf{A}_i, i=(0, 1, 2, ..., n)$. To aggregate those activation volumes, we choose to max-pool the activations along the classifier dimension, i.e.
\begin{equation}
\textbf{A}=\text{max}\{\textbf{A}_{1}, \textbf{A}_{2}, ... , \textbf{A}_{n}\}.
\end{equation}
The size of the resulted aggregated activation volume $\textbf{A}$ is still the same size as $\textbf{A}_{i}$. 

\subsection{Learning to Generate Attention Maps}
\label{sec:learn_attn}
Given $\textbf{A}$, we can obtain the attention map $\textbf{M}$ by simply max-pooling $\textbf{A}$ along the channel dimension excluding the last background category. Detailed steps are shown in Figure~\ref{fig:get_attn}. 

Since the attention map $\textbf{M}$ is generated from the output activations of local classifiers, it indicates where those classifiers pay attention to. For example, a location with higher activation value indicates that one of the local classifiers is more confident about its category prediction based on the local feature. On one hand, it is reasonable to regard this location as part of our attention region. On the other hand, before those local classifiers get well trained, we cannot guarantee the predictions within the attended region are correct predictions. Conversely, many predictions within the attended region might be incorrect in the first several training epochs. That is to say, some locations are activated as part of the attention region by false confident predictions.
However, due to the lack of detailed object annotations as supervisory signals, we face a challenge to train those local classifiers. 

\begin{figure}
\centering
\includegraphics[width=1.0\linewidth]{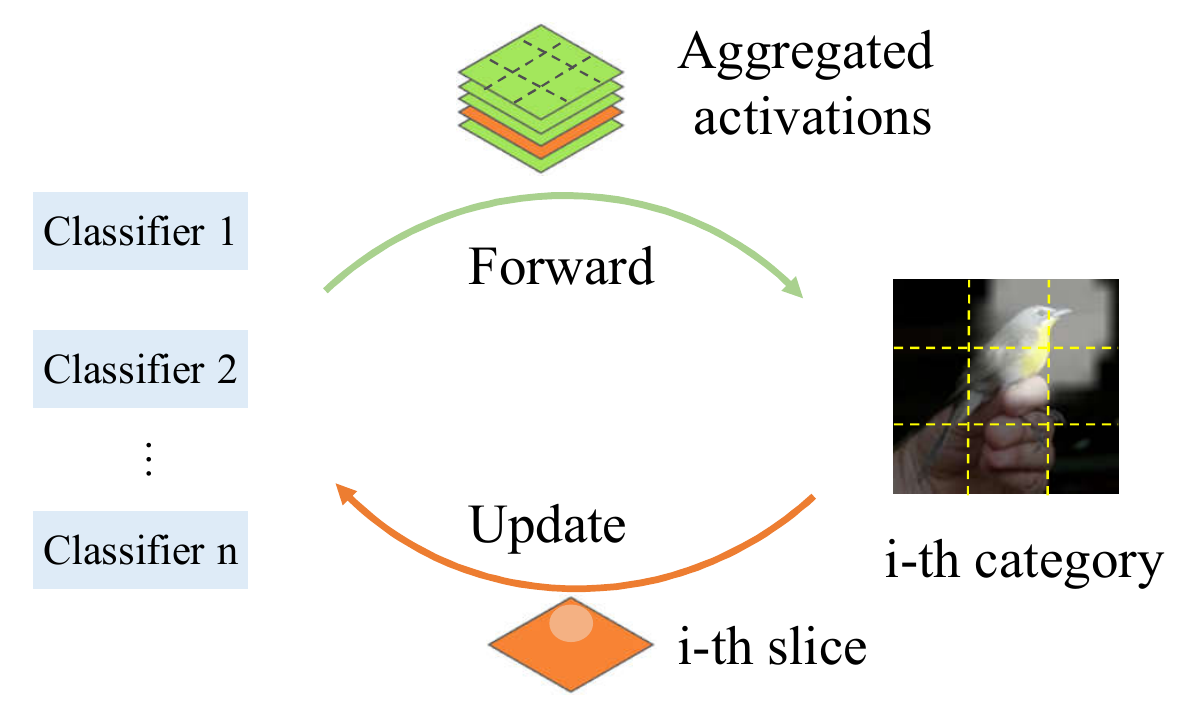}
\caption{Training local classifiers. We train local classifiers in a way similar to semantic segmentation~\cite{Long_2015_CVPR}. The binary attention map is used as the surrogate mask for the object belonging to the $i$-th category. During training, we update classifiers' weight so that the $i$-th slice in the aggregated activation volume approximates the final attention map. (Best viewed in color.)}
\label{fig:local}
\end{figure}

Inspired by semantic segmentation~\cite{Long_2015_CVPR} where each output slice corresponds to a specific category segmentation, we also treat each slice of the aggregated activations $\textbf{A}$ as an attention map corresponding to a specific category. 
Ideally, one can use accurate object masks to facilitate the training. At locations within the object mask, classifiers should make correct fine-grained category predictions. At locations beyond the object mask, classifiers should be able to label the locations as background. 
Here, we use Otsu method~\cite{otsu1979threshold} to obtain a binary object attention map $\textbf{M}_b$ from $\textbf{M}$ and regard it as the surrogate of the target fine-grained object mask which serves as the same role as the segmentation mask plays in semantic segmentation. Note that no matter whether the attended locations are activated by incorrect predictions or correct predictions we treat them equally as part of the attention region as long as their probability activations are high. The assumption is that higher probability indicates that the local feature contains abundant information which is worth our attention. Though this assumption does not always hold true. We find that it indeed works in our scenario. The learning process is shown in Figure~\ref{fig:local}.

\subsection{Fine-grained Recognition}
For the fine grained recognition task, we train our model at different image scales similar to other state-of-the-art methods~\cite{fu2017look}\cite{zhao2016diversified}. Multi-scale recognition means recursively localizing, cropping and amplifying the attended region multiple times for recognition. For each scale, we train classifiers at two levels. One is the local classifiers which are detailed in the above sections and the other is the object level classifier. 

\subsubsection{Local Level Loss Function}
As mentioned in the previous section, we use the binary attention map $\textbf{M}_b$ as the surrogate object mask to construct the loss function
\begin{equation}
\ell_{loc}=\frac{1}{WH}((1-w)\ell_1+w\ell_0), 
\label{eq:loss_local}
\end{equation}
where
\begin{equation} 
\ell_0=\sum \limits_{\{(i,j)|\textbf{M}_b(i,j)=0\}}-\text{log}p^{bk}_{(i,j)} 
\end{equation}
indicates the loss of background category classification and
\begin{equation} 
\ell_1=\sum \limits_{\{(i,j)|\textbf{M}_b(i,j)=1\}}-\text{log}p^t_{(i,j)}
\label{eq:loss_fg}
\end{equation}
indicates the loss of fine-grained category classification. $p^t_{(i,j)}$ and $p^{bk}_{(i,j)}$ are the softmax probabilities of the $t$-th category label and the background category label at location $(i, j)$ on the feature maps.
Usually the background regions are much larger than our attention regions. Therefore we adopt a weight $w$ to prevent $\ell_0$ from being too dominant. Here, $w=\frac{1}{WH}\sum \limits_{(i,j)}\textbf{M}_b(i,j)$ is the ratio of the attended area over the entire spatial area of the high level feature maps. 

\subsubsection{Object Level Loss Function}
Local classifiers only utilize local information while the object level information is ignored. Therefore, we also train a linear classifier based on object level features. Given the attention mask $\textbf{M}_b$, we spatially max-pool the weighted feature maps $\textbf{M}_bf(X)$ to a single vector as the object level features. The loss function is
\begin{equation}
\ell_{obj}=-\log p^t
\label{eq:loss_object}
\end{equation}
where $t$ is the label of the image and $p^t$ is the softmax probability of the $t$-th category.

Combing both the local level loss and the object level loss, the loss function at a single scale is 
\begin{equation}
\ell=\ell_{loc}+\ell_{obj}.
\label{eq:loss_hyb}
\end{equation}
This loss function consisting of only softmax losses enables fast implementation and easy training of the proposed model. When doing multi-scale training, our models are trained sequentially and the objective function (see Equation~\ref{eq:loss_hyb}) at each scale is optimized independently. 

\subsubsection{Multi-scale Prediction}
Once our model is well trained, we can average multi-scale predictions as the final prediction. Specifically, for each scale, the prediction is the average of both the averaged local prediction and object level prediction. The averaged local prediction is obtained by spatially average pooling the aggregated activation maps $\textbf{A}$ excluding the last background category. Note that the channel dimension of $\textbf{A}$ is $L+1$. When the background category is removed, the obtained vector naturally serves as local level prediction. The object level prediction is simply obtained from the output of the object level classifier. We average all the predicted probabilities from all scales as the final prediction.

\section{Experiments}
\subsection{Datasets}
Evaluations are performed on CUB-200-2011~\cite{welinder2010caltech}, FGVC-Aircraft~\cite{maji2013fine} and Stanford Cars~\cite{krause20133d}. A brief description of the three datasets including the train-test split and the number of categories is given in Table~\ref{tb_exp:datasets}.

\setlength{\tabcolsep}{4pt}
\begin{table}
\begin{center}
\caption{A brief description of the datasets for evaluation}
\label{tb_exp:datasets}
\begin{tabular}{cccc}
\hline\noalign{\smallskip}
Datasets  & \# Category & \# Training & \# Testing\\
\noalign{\smallskip}
\hline
\noalign{\smallskip}
CUB-200-2011		 & 200 	& 5,994	& 5,794\\
FGVC-Aircraft		 & 100	& 6,667 & 3,333\\
Stanford Cars			 & 196	& 8,144 & 8,041\\
\hline
\end{tabular}
\end{center}
\end{table}
\setlength{\tabcolsep}{1.4pt}

\subsection{Implementation Details}
ResNet50~\cite{he2016deep} pre-trained on the ImageNet classification dataset is adopted as the backbone network. The convolutional feature maps before the last pooling layer are extracted as the high level feature representation. All the local classifiers and the object level classifier are linear classifiers. For each local classifier, it is implemented as a $(L+1)\times 1 \times 1$ convolution kernel so that our model can be trained in an end-to-end way. For all images in the datasets, we resize their short edges to 448 as~\cite{zheng2017learning,fu2017look}.
The number of local classifiers is 16 and the fine tuning learning rate is 1e-4. Each model is trained for 40 epochs. 




\setlength{\tabcolsep}{4pt}
\begin{table}
\begin{center}
\caption{Comparison with state-of-the-art methods on CUB-200-2011 dataset. $^*$ indicates the model is trained on the multi-scale images generated by our model.}
\label{tb_exp:cub}
\begin{tabular}{ccc}
\hline\noalign{\smallskip}
CUB-200-2011  & Structured Annotation & Accuracy(\%)\\
\noalign{\smallskip}
\hline
\noalign{\smallskip}
PA-CNN~\cite{krause2015fine}				 & Yes	& 82.8\\
MG-CNN~\cite{wang2015multiple}				 & Yes	& 83.0 	\\
Mask-CNN~\cite{wei2018mask}				 & Yes	& 87.3 	\\
HSNet~\cite{Lam_2017_CVPR} & Yes	& 87.5 	\\
B-CNN~\cite{lin2015bilinear}				 & No	& 84.1 	\\
ST-CNN ~\cite{jaderberg2015spatial}				 & No	& 84.1 	\\
PDFR~\cite{zhang2016picking}				 & No	& 84.5 	\\
RA-CNN~\cite{fu2017look}				 & No	& 85.3 	\\
MA-CNN~\cite{zheng2017learning}				 & No	& 86.5 	\\
G$^2$DeNet~\cite{Wang_2017_CVPR}				 & No	& 87.1 	\\
\hline
ResNet50(single scale)				 & -	& 84.0 	\\
ResNet50(multi-scale)$^*$				 & -	& 85.1 	\\
ours				 & No	& \textbf{87.9} 	\\
\hline
\end{tabular}
\end{center}
\end{table}
\setlength{\tabcolsep}{1.4pt}

\setlength{\tabcolsep}{4pt}
\begin{table}
\begin{center}
\caption{Comparison with state-of-the-art methods on Stanford Cars dataset. $^*$ indicates the model is trained on the multi-scale images generated by our model.}
\label{tb_exp:car}
\begin{tabular}{ccc}
\hline\noalign{\smallskip}
Stanford Cars  & Structured Annotation & Accuracy(\%)\\
\noalign{\smallskip}
\hline
\noalign{\smallskip}
R-CNN~\cite{branson2014bird}	 		 & Yes 	& 88.4\\
MDTP~\cite{wang2016mining}				& Yes	& 91.3 	\\
PA-CNN~\cite{krause2015fine}				 & Yes	& 92.8\\
HSNet~\cite{Lam_2017_CVPR} & Yes	& 93.9 	\\
B-CNN~\cite{lin2015bilinear}				 & No	& 91.3 	\\
RA-CNN~\cite{fu2017look}				 & No	& 92.5 	\\
G$^2$DeNet~\cite{Wang_2017_CVPR}				 & No	& 92.5 	\\
MA-CNN~\cite{zheng2017learning}				 & No	& 92.8 	\\
\hline
ResNet50(single scale)				 & -	& 91.2 	\\
ResNet50(multi-scale)$^*$				 & -	& 92.0 	\\
ours				 & No	& \textbf{94.1} 	\\
\hline
\end{tabular}
\end{center}
\end{table}
\setlength{\tabcolsep}{1.4pt}

\setlength{\tabcolsep}{4pt}
\begin{table}
\begin{center}
\caption{Comparison with state-of-the-art methods on FGVC-Aircraft dataset. $^*$ indicates the model is trained on the multi-scale images generated by our model.}
\label{tb_exp:dog}
\begin{tabular}{ccc}
\hline\noalign{\smallskip}
FGVC-Aircraft  & Structured Annotation & Accuracy(\%)\\
\noalign{\smallskip}
\hline
\noalign{\smallskip}
MG-CNN~\cite{wang2015multiple}	& Yes	& 86.6 \\
MDTP~\cite{wang2016mining}		& Yes	& 88.4 	\\
FV-CNN~\cite{gosselin2014revisiting}	& No	& 81.5 	\\
B-CNN~\cite{lin2015bilinear}	& No	& 84.1 	\\
RA-CNN~\cite{fu2017look}				 & No	& 88.2 	\\
G$^2$DeNet~\cite{Wang_2017_CVPR}				 & No	& 89.0 	\\
MA-CNN~\cite{zheng2017learning}	& No	& 89.9 	\\
\hline
ResNet50(single scale)				 & -	& 87.9 	\\
ResNet50(multi-scale)$^*$				 & -	& 89.1 	\\
ours	& No	& \textbf{92.1} 	\\
\hline
\end{tabular}
\end{center}
\end{table}
\setlength{\tabcolsep}{1.4pt}

\subsection{Evaluation of the Learned Attention}
Before the presentation of attention evaluation, we firstly introduce our baseline models, \ie  a single scale ResNet50 model~\cite{he2016deep} and a multi-scale ResNet50 model. For both models, we spatially max-pool the final convolutional feature maps to obtain an entire image-level feature vector for classification. The single scale ResNet50 is trained on the original images in the datasets while the mutli-scale ResNet50 is trained on multi-scale images. Note that ResNet50 does not directly generate attention regions to focus on. Thus, we use the multi-scale images generated by our model to train the multi-scale ResNet50. Under this setting, both our model and multi-scale ResNet50 will be trained on the same multi-scale images with more and more focused regions. For multi-scale ResNet-50,  we average predictions from three scales as the final prediction. We show the recognition results on three datasets from Table~\ref{tb_exp:cub} to Table~\ref{tb_exp:dog} along with other state-of-the-art methods.

From the results, we can find that our model outperforms the baseline multi-scale ResNet50 by more than 2\%. Since both models are trained on the same multi-scale images, we argue that the improvement is brought by our generated attention which facilitates the training of both local classifiers and the object level classifier at each scale. We also note that the recognition accuracy of the multi-scale ResNet50 model is higher than the single scale ResNet50 model. This can be explained by the fact that the multi-scale ResNet50 model is trained on multi-scale images generated by our model which enables the model to capture more and more discriminative features. It also demonstrates the effectiveness of the generated attention. Even the attention generation network is trained within our model, the generated attention maps are also helpful for other models to improve recognition accuracy.

\subsection{Comparison with State-of-the-Art Methods}
We show the recognition result comparisons on three datasets from Table~\ref{tb_exp:cub} to Table~\ref{tb_exp:dog}.
The classification accuracy of our model surpasses the recent models on all three benchmark datasets. The model that is most similar to our model is RA-CNN~\cite{fu2017look}. We find our model surpasses RA-CNN by around 2\%. We also notice that our model even outperforms most recent Mask-CNN~\cite{wei2018mask} and HSNet~\cite{Lam_2017_CVPR} which use extra annotations on CUB-200-2011 dataset and Standford Cars dataset. 

It is worth mentioning that the objective function of our model contains only the commonly used softmax classification loss while the objective function of RA-CNN contains dedicated ranking loss besides the classification loss. It shows that even our model uses a much simpler objective function, the recognition accuracy is higher than that of other state-of-the-art methods.
The recognition results on three benchmark datasets demonstrate that our model is an effective architecture for fine-grained recognition. 


\begin{figure}
\centering
\includegraphics[width=0.98\linewidth]{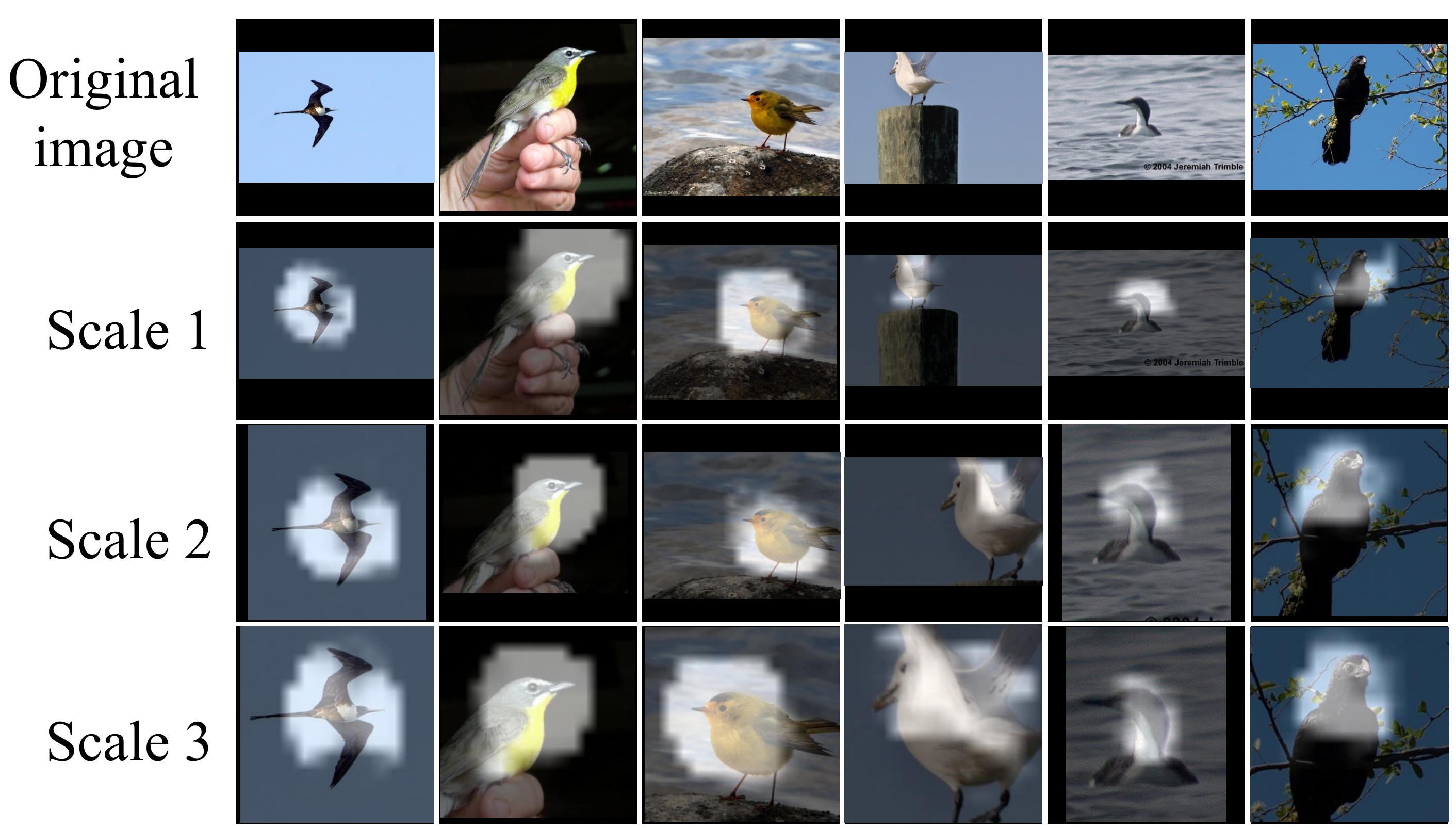}
\caption{Bird examples of multi-scale attentions. Images are overlaid by attention maps. Brighter regions indicate higher possibilities of being discriminative. (Best viewed in color.)}
\label{fig:exp:attn_cub}
\end{figure}
\begin{figure}
\centering
\includegraphics[width=0.98\linewidth]{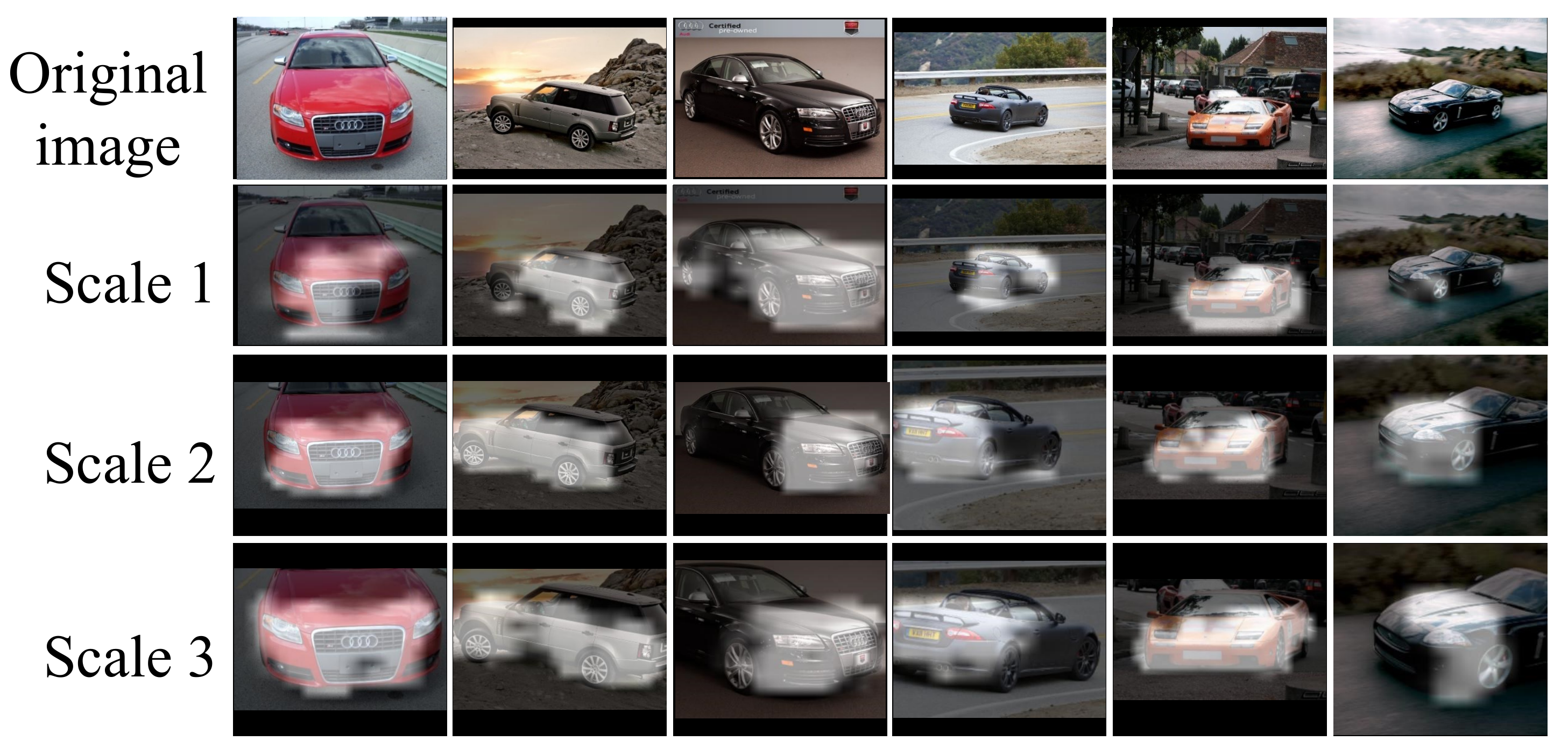}
\caption{Car examples of multi-scale attentions. Images are overlaid by attention maps. Brighter regions indicate higher possibilities of being discriminative. (Best viewed in color.)}
\label{fig:exp:attn_car}
\end{figure}
\begin{figure}
\centering
\includegraphics[width=0.98\linewidth]{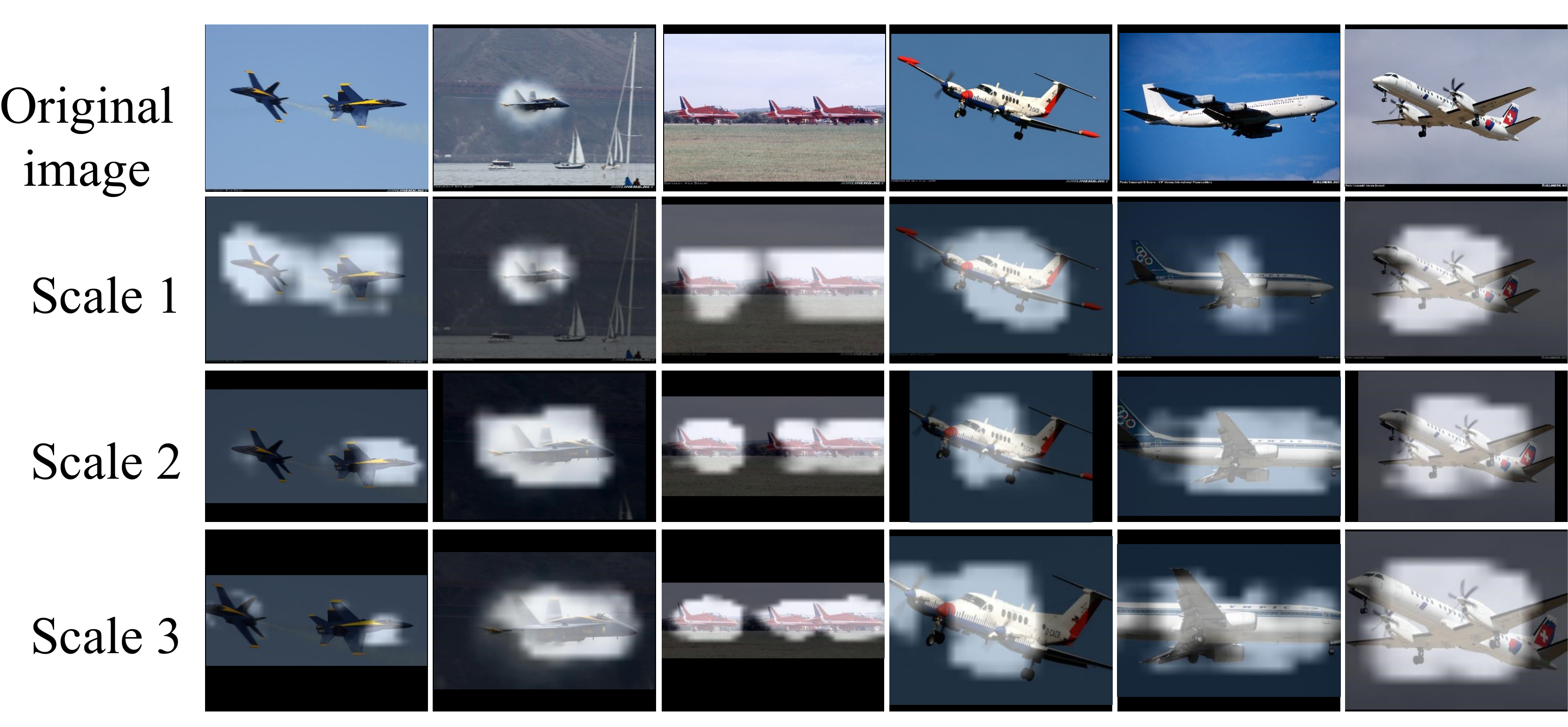}
\caption{Aircraft examples of multi-scale attentions. Images are overlaid by attention maps. Brighter regions indicate higher possibilities of being discriminative. (Best viewed in color.)}
\label{fig:exp:attn_air}
\end{figure}
\begin{figure}
\centering
\includegraphics[width=0.98\linewidth]{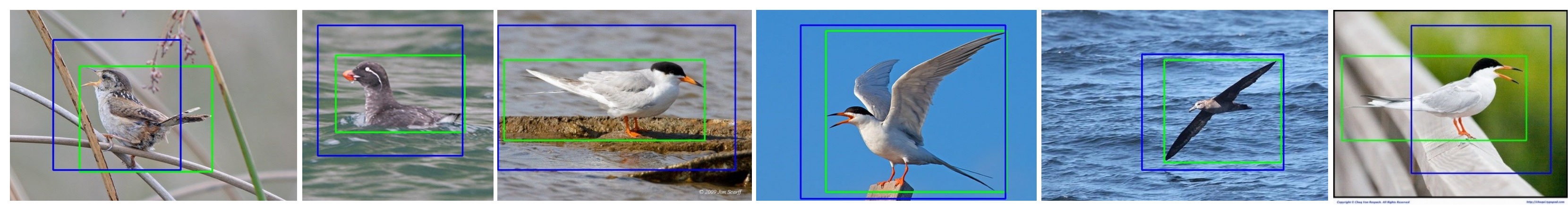}
\caption{Comparison of ground truth bounding boxes (green) and bounding boxes calculated from attended regions (blue). (Best viewed in color.)}
\label{fig:exp:attn_cub_bbox}
\end{figure}
\subsection{Attention Visualization}
From Figure~\ref{fig:exp:attn_cub} to Figure~\ref{fig:exp:attn_air}, examples of the learned attention are illustrated for the purpose of intuitive understanding.
It is obvious that our model can pay attention to the discriminative object regions and exclude the cluttered background. Unlike the model in ~\cite{fu2017look} which  
generates a single attended bounding box, our model generates attention masks. In some cases where there are multiple objects, our model can simutaneously localize multiple objects which is shown in the first and third column in Figure~\ref{fig:exp:attn_air}. The loss function in Equation~\ref{eq:loss_fg} encourages our model to generate a slightly larger attention regions than the attention regions in~\cite{fu2017look}. Nevertheless, we can still find that the attention regions from the second and the last scales are much more focused compared to the attention generated form the first scale.

Since our model can localize object regions, it is interesting to see the relationship between the bounding boxes of the attended regions and those of the ground truth objects. We choose binary attention maps from the first scale as the attended object regions since our model focuses more on the whole object at this scale.
Here, this experiment is conducted on CUB-200-2011 dataset for an intuitive illustration. Some calculated bounding boxes are shown in Figure~\ref{fig:exp:attn_cub_bbox}. It can be seen that the  bounding boxes of the attended regions approximate the ground truth bounding boxes. The averaged IOU between the bounding boxes of the attended regions and the ground truth bounding boxes is 0.54. The IOU is promising since the obtained attended regions are optimized for fine-grained classification which is quite different from traditional object detection tasks. Therefore, we do not compare our results to other object detection methods.

\subsection{Ablation Study}

\subsubsection{Effect of Local Level Loss and Object Level Loss}
One might argue that a single object level loss is sufficient for the recognition task.
To investigate the effectiveness of those two level losses, we train one model with only the local level loss and another model with only the object level loss. The results are shown in Table~\ref{tb_exp:single_loss}. 
Compared to the recognition accuracy from Table~\ref{tb_exp:cub} to Table~\ref{tb_exp:dog}, we notice an obvious accuracy drop when the model is trained with only a single loss. It can be explained that when the model is trained under the hybrid loss, it can get more supervised signals from both local level and object level information, leading to more efficient learning. The local loss can also be regarded as a regularization term that reduces the risk of the object level classifier over-fitting the dataset.

\setlength{\tabcolsep}{4pt}
\begin{table}
\begin{center}
\caption{Recognition accuracy (\%) when our model is trained with only the local level loss and trained with only the object level loss on three benchmark datasets.}
\label{tb_exp:single_loss}
\begin{tabular}{c|c|c|c}
\hline
           & CUB-2011-200 & Stanford-Cars & FGVC-Aircraft \\ \hline
Acc$_{loc}$ & 83.9         & 88.0          & 86.0          \\ \hline
Acc$_{obj}$ & 85.6         & 92.2          & 89.3         \\ \hline
\end{tabular}
\end{center}
\end{table}
\setlength{\tabcolsep}{1.4pt}

\setlength{\tabcolsep}{4pt}
\begin{table*}
\begin{center}
\caption{Recognition accuracy (\%) from multi-scale models. The numbers in the second row indicate the first, the second and the third scale while \lq\lq ms \rq\rq means multi-scale. \lq\lq Acc$_{loc}$ \rq\rq, \lq\lq Acc$_{obj}$ \rq\rq and \lq\lq Acc$_{avg}$ \rq\rq stand for the recognition accuracy of the averaged local predictions, the object level prediction and the final averaged prediction of both local level predictions and object level prediction, respectively.}
\label{tb_exp:local_global_hyb}
\begin{tabular}{c|c|c|c|c|c|c|c|c|c|c|c|c}
\hline
dataset & \multicolumn{4}{c|}{CUB-2011-200}      & \multicolumn{4}{c|}{Stanford-Cars}     & \multicolumn{4}{c}{FGVC-Aircraft}     \\ \hline
scale      & 1 & 2 & 3 & ms & 1 & 2 & 3 & ms & 1 & 2 & 3 & ms \\ \hline
Acc$_{loc}$        & 84.8   & 85.2   & 85.4   & 86.8        & 92.7   & 92.9   & 92.1   & 93.4        & 90.0   & 88.6   & 87.4   & 90.5        \\ \hline
Acc$_{obj}$       & 85.3   & 85.4   & 85.9   & 87.7        & 92.8   & 93.4   & 92.4   & 93.9        & 90.5   & 89.9   & 88.6   & 91.5        \\ \hline
Acc$_{avg}$		 & 85.5   & 85.4   & 86.1   & \textbf{87.9}        & 92.9   & 93.6   & 92.7   & \textbf{94.1}        & 90.7   &  90.1  & 88.6   & \textbf{92.1}        \\ \hline
\end{tabular}
\end{center}
\end{table*}
\setlength{\tabcolsep}{1.4pt}

\subsubsection{Effect of Combining Local Level Predictions and Object Level Prediction}
At each scale, the final prediction is the averaged prediction of both averaged local predictions and the object level prediction. We show the recognition accuracy of the averaged local predictions and the object level prediction separately in each row of Table~\ref{tb_exp:local_global_hyb}.
Note that unlike previous experiment, Acc$_{local}$ and Acc$_{obj}$ in this section are obtained from models trained under both the local level loss and the object level loss. 
We find that the recognition accuracy of averaged local predictions is slightly lower than that of the object level prediction. It is reasonable because the features used in dense local activations are local features which may lack sufficient discriminative information for recognition. In contrast, the feature for the object level prediction is the aggregation of a  feature set within the attended region which contains more object information.
By comparing Acc$_{loc}$, Acc$_{obj}$ and Acc$_{avg}$, we observe that combing both local level prediction and object level prediction yields higher recognition accuracy at most scales.

\subsubsection{Effect of Multi-scale Recognition}
To illustrate the effectiveness of multi-scale  recognition, we report the recognition accuracy from each scale in Table~\ref{tb_exp:local_global_hyb}.  
Comparing the accuracy in the multi-scale column to those in single scale columns, we find that the accuracy in the multi-scale column is higher than any of a single scale. It demonstrates that the multi-scale ensemble indeed helps improve recognition accuracy.

\subsubsection{Effect of the Number of Local Classifiers}
\begin{figure}
\centering
\includegraphics[width=1.0\linewidth]{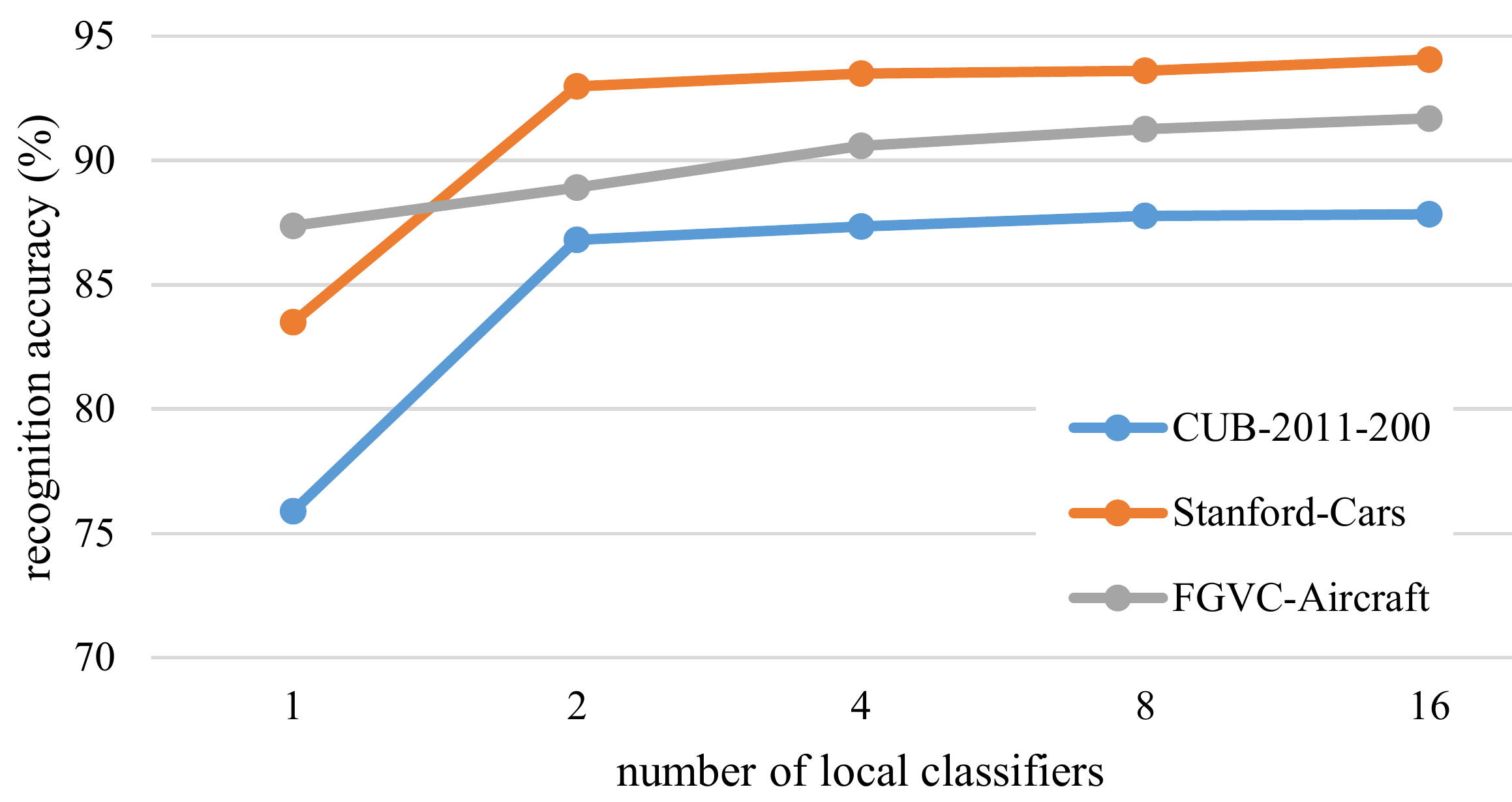}
\caption{Recognition accuracy comparison of our models with different number of local classifiers. (Best viewed in color.)}
\label{fig:exp:num_clfs}
\end{figure}

For dense local activations, we use $n$ local classifiers. In this section, we investigate the effect of the number of local classifiers. We retrain our model with 1, 2, 4, 8, and 16 local classifiers, respectively. The recognition results are shown in Figure~\ref{fig:exp:num_clfs}. From the figure, we can find that the recognition accuracy firstly increases and then statures as the number of classifiers increases. Surprisingly, when there are two local classifiers, the recognition accuracy increases by a large margin compare to the that of training using only one classifier. The reason is that images are high dimensional data such that using a single local classifier is insufficient to make correct recognition at all locations. By adding more local classifiers, we increase the capacity of the local model so that it can perform better in terms of recognition accuracy.
\section{Conclusion}
In this paper, we propose to generate attention regions based on the activations of local classifiers for fine-grained recognition. The proposed model can be trained with only image level labels and softmax loss which makes it simple and easy to implement. Evaluation on three benchmark datasets demonstrates the superior performance of our model. 

{\small
\bibliographystyle{ieee}
\bibliography{egbib}

\begin{thebibliography}{10}\itemsep=-1pt

\bibitem{branson2014bird}
S.~Branson, G.~Van~Horn, P.~Perona, and S.~Belongie.
\newblock Bird species categorization using pose normalized deep convolutional
  nets.
\newblock In {\em Proceedings of the British Machine Vision Conference}. BMVA
  Press, 2014.

\bibitem{chen2016attention}
L.-C. Chen, Y.~Yang, J.~Wang, W.~Xu, and A.~L. Yuille.
\newblock Attention to scale: Scale-aware semantic image segmentation.
\newblock In {\em Conference on Computer Vision and Pattern Recognition}, pages
  3640--3649, 2016.

\bibitem{connor2004visual}
C.~E. Connor, H.~E. Egeth, and S.~Yantis.
\newblock Visual attention: bottom-up versus top-down.
\newblock {\em Current biology}, 14(19):R850--R852, 2004.

\bibitem{Cui_2018_CVPR}
Y.~Cui, Y.~Song, C.~Sun, A.~Howard, and S.~Belongie.
\newblock Large scale fine-grained categorization and domain-specific transfer
  learning.
\newblock In {\em The IEEE Conference on Computer Vision and Pattern
  Recognition (CVPR)}, June 2018.

\bibitem{fu2017look}
J.~Fu, H.~Zheng, and T.~Mei.
\newblock Look closer to see better: Recurrent attention convolutional neural
  network for fine-grained image recognition.
\newblock In {\em Conference on Computer Vision and Pattern Recognition}, 2017.

\bibitem{gosselin2014revisiting}
P.-H. Gosselin, N.~Murray, H.~J{\'e}gou, and F.~Perronnin.
\newblock Revisiting the fisher vector for fine-grained classification.
\newblock {\em Pattern recognition letters}, 49:92--98, 2014.

\bibitem{gregor2015draw}
K.~Gregor, I.~Danihelka, A.~Graves, D.~Rezende, and D.~Wierstra.
\newblock Draw: A recurrent neural network for image generation.
\newblock In {\em International Conference on Machine Learning}, pages
  1462--1471, 2015.

\bibitem{he2016deep}
K.~He, X.~Zhang, S.~Ren, and J.~Sun.
\newblock Deep residual learning for image recognition.
\newblock In {\em Conference on Computer Vision and Pattern Recognition}, pages
  770--778, 2016.

\bibitem{huang2016part}
S.~Huang, Z.~Xu, D.~Tao, and Y.~Zhang.
\newblock Part-stacked cnn for fine-grained visual categorization.
\newblock In {\em Conference on Computer Vision and Pattern Recognition}, pages
  1173--1182, 2016.

\bibitem{jaderberg2015spatial}
M.~Jaderberg, K.~Simonyan, A.~Zisserman, et~al.
\newblock Spatial transformer networks.
\newblock In {\em Advances in neural information processing systems}, pages
  2017--2025, 2015.

\bibitem{krause2014learning}
J.~Krause, T.~Gebru, J.~Deng, L.-J. Li, and L.~Fei-Fei.
\newblock Learning features and parts for fine-grained recognition.
\newblock In {\em International Conference on Pattern Recognition}, pages
  26--33. IEEE, 2014.

\bibitem{krause2015fine}
J.~Krause, H.~Jin, J.~Yang, and L.~Fei-Fei.
\newblock Fine-grained recognition without part annotations.
\newblock In {\em Conference on Computer Vision and Pattern Recognition}, pages
  5546--5555. IEEE, 2015.

\bibitem{krause20133d}
J.~Krause, M.~Stark, J.~Deng, and L.~Fei-Fei.
\newblock 3d object representations for fine-grained categorization.
\newblock In {\em International Conference on Computer Vision Workshops}, pages
  554--561. IEEE, 2013.

\bibitem{Lam_2017_CVPR}
M.~Lam, B.~Mahasseni, and S.~Todorovic.
\newblock Fine-grained recognition as hsnet search for informative image parts.
\newblock In {\em The IEEE Conference on Computer Vision and Pattern
  Recognition (CVPR)}, July 2017.

\bibitem{lin2015deep}
D.~Lin, X.~Shen, C.~Lu, and J.~Jia.
\newblock Deep lac: Deep localization, alignment and classification for
  fine-grained recognition.
\newblock In {\em Conference on Computer Vision and Pattern Recognition}, pages
  1666--1674. IEEE, 2015.

\bibitem{lin2015bilinear}
T.-Y. Lin, A.~RoyChowdhury, and S.~Maji.
\newblock Bilinear cnn models for fine-grained visual recognition.
\newblock In {\em International Conference on Computer Vision}, pages
  1449--1457, 2015.

\bibitem{liu2017localizing}
X.~Liu, J.~Wang, S.~Wen, E.~Ding, and Y.~Lin.
\newblock Localizing by describing: Attribute-guided attention localization for
  fine-grained recognition.
\newblock In {\em AAAI}, pages 4190--4196, 2017.

\bibitem{Long_2015_CVPR}
J.~Long, E.~Shelhamer, and T.~Darrell.
\newblock Fully convolutional networks for semantic segmentation.
\newblock In {\em The IEEE Conference on Computer Vision and Pattern
  Recognition (CVPR)}, June 2015.

\bibitem{maji2013fine}
S.~Maji, E.~Rahtu, J.~Kannala, M.~Blaschko, and A.~Vedaldi.
\newblock Fine-grained visual classification of aircraft.
\newblock {\em arXiv preprint arXiv:1306.5151}, 2013.

\bibitem{Niu_2018_CVPR}
L.~Niu, A.~Veeraraghavan, and A.~Sabharwal.
\newblock Webly supervised learning meets zero-shot learning: A hybrid approach
  for fine-grained classification.
\newblock In {\em The IEEE Conference on Computer Vision and Pattern
  Recognition (CVPR)}, June 2018.

\bibitem{otsu1979threshold}
N.~Otsu.
\newblock A threshold selection method from gray-level histograms.
\newblock {\em IEEE Transactions on Systems, Man, and Cybernetics},
  9(1):62--66, 1979.

\bibitem{44891}
P.~Sermanet, A.~Frome, and E.~Real.
\newblock Attention for fine-grained categorization.
\newblock In {\em International Conference on Learning Representations
  Workshop}, 2015.

\bibitem{simon2015neural}
M.~Simon and E.~Rodner.
\newblock Neural activation constellations: Unsupervised part model discovery
  with convolutional networks.
\newblock In {\em International Conference on Computer Vision}, pages
  1143--1151, 2015.

\bibitem{wang2015multiple}
D.~Wang, Z.~Shen, J.~Shao, W.~Zhang, X.~Xue, and Z.~Zhang.
\newblock Multiple granularity descriptors for fine-grained categorization.
\newblock In {\em International Conference on Computer Vision}, pages
  2399--2406. IEEE, 2015.

\bibitem{wang2017residual}
F.~Wang, M.~Jiang, C.~Qian, S.~Yang, C.~Li, H.~Zhang, X.~Wang, and X.~Tang.
\newblock Residual attention network for image classification.
\newblock In {\em Conference on Computer Vision and Pattern Recognition}, pages
  3156--3164, 2017.

\bibitem{wang2017multi}
P.~Wang, L.~Liu, C.~Shen, Z.~Huang, A.~van~den Hengel, and H.~Tao~Shen.
\newblock Multi-attention network for one shot learning.
\newblock In {\em Conference on Computer Vision and Pattern Recognition}, pages
  2721--2729, 2017.

\bibitem{Wang_2017_CVPR}
Q.~Wang, P.~Li, and L.~Zhang.
\newblock G2denet: Global gaussian distribution embedding network and its
  application to visual recognition.
\newblock In {\em The IEEE Conference on Computer Vision and Pattern
  Recognition (CVPR)}, July 2017.

\bibitem{wang2016mining}
Y.~Wang, J.~Choi, V.~I. Morariu, and L.~S. Davis.
\newblock Mining discriminative triplets of patches for fine-grained
  classification.
\newblock In {\em Conference on Computer Vision and Pattern Recognition}, pages
  1163--1172. IEEE, 2016.

\bibitem{wei2018mask}
X.-S. Wei, C.-W. Xie, J.~Wu, and C.~Shen.
\newblock Mask-cnn: Localizing parts and selecting descriptors for fine-grained
  bird species categorization.
\newblock {\em Pattern Recognition}, 76:704--714, 2018.

\bibitem{welinder2010caltech}
P.~Welinder, S.~Branson, T.~Mita, C.~Wah, F.~Schroff, S.~Belongie, and
  P.~Perona.
\newblock Caltech-ucsd birds 200.
\newblock 2010.

\bibitem{xiao2015application}
T.~Xiao, Y.~Xu, K.~Yang, J.~Zhang, Y.~Peng, and Z.~Zhang.
\newblock The application of two-level attention models in deep convolutional
  neural network for fine-grained image classification.
\newblock In {\em Conference on Computer Vision and Pattern Recognition}, pages
  842--850. IEEE, 2015.

\bibitem{xu2016ask}
H.~Xu and K.~Saenko.
\newblock Ask, attend and answer: Exploring question-guided spatial attention
  for visual question answering.
\newblock In {\em European Conference on Computer Vision}, pages 451--466.
  Springer, 2016.

\bibitem{xu2015show}
K.~Xu, J.~Ba, R.~Kiros, K.~Cho, A.~Courville, R.~Salakhudinov, R.~Zemel, and
  Y.~Bengio.
\newblock Show, attend and tell: Neural image caption generation with visual
  attention.
\newblock In {\em International Conference on Machine Learning}, pages
  2048--2057, 2015.

\bibitem{yang2016stacked}
Z.~Yang, X.~He, J.~Gao, L.~Deng, and A.~Smola.
\newblock Stacked attention networks for image question answering.
\newblock In {\em Conference on Computer Vision and Pattern Recognition}, pages
  21--29, 2016.

\bibitem{zhang2016spda}
H.~Zhang, T.~Xu, M.~Elhoseiny, X.~Huang, S.~Zhang, A.~Elgammal, and D.~Metaxas.
\newblock Spda-cnn: Unifying semantic part detection and abstraction for
  fine-grained recognition.
\newblock In {\em Conference on Computer Vision and Pattern Recognition}, pages
  1143--1152, 2016.

\bibitem{zhang2016top}
J.~Zhang, Z.~Lin, J.~Brandt, X.~Shen, and S.~Sclaroff.
\newblock Top-down neural attention by excitation backprop.
\newblock In {\em European Conference on Computer Vision}, pages 543--559.
  Springer, 2016.

\bibitem{zhang2014part}
N.~Zhang, J.~Donahue, R.~Girshick, and T.~Darrell.
\newblock Part-based r-cnns for fine-grained category detection.
\newblock In {\em European Conference on Computer Vision}, pages 834--849.
  Springer, 2014.

\bibitem{zhang2016picking}
X.~Zhang, H.~Xiong, W.~Zhou, W.~Lin, and Q.~Tian.
\newblock Picking deep filter responses for fine-grained image recognition.
\newblock In {\em Conference on Computer Vision and Pattern Recognition}, pages
  1134--1142, 2016.

\bibitem{zhao2016diversified}
B.~Zhao, X.~Wu, J.~Feng, Q.~Peng, and S.~Yan.
\newblock Diversified visual attention networks for fine-grained object
  classification.
\newblock {\em IEEE Transactions on Multimedia}, 19(6):1245--1256, 2017.

\bibitem{zheng2017learning}
H.~Zheng, J.~Fu, T.~Mei, and J.~Luo.
\newblock Learning multi-attention convolutional neural network for
  fine-grained image recognition.
\newblock In {\em International Conference on Computer Vision}, 2017.

\bibitem{zhou2016learning}
B.~Zhou, A.~Khosla, A.~Lapedriza, A.~Oliva, and A.~Torralba.
\newblock Learning deep features for discriminative localization.
\newblock In {\em Computer Vision and Pattern Recognition (CVPR), 2016 IEEE
  Conference on}, pages 2921--2929. IEEE, 2016.

\end{thebibliography}
}

\end{document}